\documentclass[10pt,journal,compsoc]{IEEEtran}
\usepackage{subfigure}
\usepackage{graphicx}
\usepackage{multirow}
\usepackage{algorithm}
\usepackage{algorithmic}
\usepackage{url}
\usepackage{balance}
\usepackage{amsmath,amssymb,amsfonts}

\ifCLASSOPTIONcompsoc
  \usepackage[nocompress]{cite}
\else
  \usepackage{cite}
\fi
\ifCLASSINFOpdf
\else
\fi
\begin{document}
%
\title{Modeling Dyadic Conversations \\for Personality Inference}
%
%
%
%

\author{Qiang Liu
}

%
%

\markboth{Journal of \LaTeX\ Class Files,~Vol.~14, No.~8, August~2015}%
{Shell \MakeLowercase{\textit{et al.}}: Bare Demo of IEEEtran.cls for Computer Society Journals}
\IEEEtitleabstractindextext{%
\begin{abstract}
Nowadays, automatical personality inference is drawing extensive attention from both academia and industry. Conventional methods are mainly based on user generated contents, e.g., profiles, likes, and texts of an individual, on social media, which are actually not very reliable. In contrast, dyadic conversations between individuals can not only capture how one expresses oneself, but also reflect how one reacts to different situations. Rich contextual information in dyadic conversation can explain an individual's response during his or her conversation. In this paper, we propose a novel augmented Gated Recurrent Unit (GRU) model for learning unsupervised Personal Conversational Embeddings (PCE) based on dyadic conversations between individuals. We adjust the formulation of each layer of a conventional GRU with sequence to sequence learning and personal information of both sides of the conversation. Based on the learned PCE, we can infer the personality of each individual. We conduct experiments on the Movie Script dataset, which is collected from conversations between characters in movie scripts. We find that modeling dyadic conversations between individuals can significantly improve personality inference accuracy. Experimental results illustrate the successful performance of our proposed method.
\end{abstract}

}

\maketitle

\IEEEdisplaynontitleabstractindextext

%
\IEEEpeerreviewmaketitle

\IEEEraisesectionheading{\section{Introduction}\label{sec:introduction}}

%
%
%
%
\IEEEPARstart{R}{esearch} in psychology has suggested that behaviors and preferences of individuals can be explained to a great extent by underlying psychological constructs: personality traits \cite{farnadi2016computational,ozer2006personality}. Understanding an individual's personality can give us hints to make predictions about his or her preferences across different contexts and environments, which can be applied to enhance recommender systems \cite{lambiotte2014tracking} and advertisement targeting \cite{chen2015making}. Some research shows that personality can significant affect users' choices of social relationships \cite{asendorpf1998personality}, products \cite{kosinski2013private}, entertainment \cite{cantador2013relating} and websites \cite{kosinski2014manifestations}. The most widely-used measurement of personality is the Big Five model, which contains five personality traits: Extraversion, Agreeableness, Conscientiousness, Neuroticism and Openness \cite{john1999big}. Detailed explanations of these five personality traits can be found in Table \ref{tab:big5}.

Traditional methods for personality inference require individuals to answer hundreds of questions formulated based on psychological research \cite{john1999big}. This requires the collaboration of individuals and is very time consuming, which makes such methods hard to be applied to a large number of users on the web. Recently, extensive research has shown that user generated contents on the web can be utilized to automatically infer personality \cite{farnadi2016computational}. Some works \cite{kosinski2013private,youyou2015computer} make prediction of user personality based on their likes on Facebook, and show this technology to be practical. Some works \cite{park2015automatic,schwartz2013personality} infer personality according to textual contents posted on social media, and show language usage is a key factor for inferring personality. There are also some works \cite{liu2016analyzing} that models pictures on social media for personality inference.

However, user generated contents are not reliable enough for personality inference. Because user generated contents can only capture how you express yourself, which is usually in a way that you want others know about you. For example, users tend to show their good and friendly sides on social media. In contrast, conversation records between individuals can not only capture how you express yourself, but also reflect how you react to different situations and individuals. This is usually in a more natural way that you can not help but to say something in certain situations. Someone showing the friendly side on social media may be unfriendly in the daily life and conversations. And someone that is not active on social media may also prefer to talk a lot with his or her friends and parents. Contextual information in conversation records can provide reasons and explanations for an individual's response. For example, a good tempered person would be angry when someone says something rude to him or her. But this does not make this person bad tempered. Meanwhile, there are a growing number of applications associated with dyadic conversation data, such as online chatbots, e.g., replying and commenting on social media, e.g., Twitter\footnote{https://twitter.com/} and Weibo\footnote{http://weibo.com}, and email replying \cite{kooti2015evolution}. Accordingly, it is vital to model dyadic conversations between individuals for personality inference.

\begin{table}[tb]
  \centering
  \caption{Explanations of the Big Five personality model.}
    \begin{tabular}{cc}
    \hline
    Personality & Explanation \\
    \hline
    \multirow{2}[0]{*}{Extraversion} & A tendency to seek stimulation \\
          & in the company of others. \\
    \multirow{2}[0]{*}{Agreeableness} & A tendency to be compassionate \\
          & and cooperative towards others. \\
    \multirow{2}[0]{*}{Conscientiousness} & A tendency to act in an organized \\
          & or spontaneous way. \\
    \multirow{2}[0]{*}{Neuroticism} & A tendency to have sensitive \\
          & emotions to the environment. \\
    \multirow{2}[0]{*}{Openness} & A tendency to be open to \\
          & experiencing a variety of activities. \\
    \hline
    \end{tabular}%
  \label{tab:big5}%
\end{table}%

In this work, we learn unsupervised \textbf{Personal Conversational Embeddings (PCE)} based on dyadic conversations between individuals and utilize the learned PCE of each individual for personality inference. Recently, Recurrent Neural Networks (RNN) have been widely applied and achieved successful performance in modeling sentences. Considering the vanishing or exploding gradients problem \cite{bengio1994learning} in conventional RNN architecture, and the high computational cost in the Long Short-Term Memory (LSTM) unit \cite{hochreiter1997long}, we select the Gated Recurrent Unit (GRU) \cite{chung2014empirical} for learning sentence embeddings. However, GRU can only model the meaning of a single sentence or paragraph, which will miss the following three important contextual factors in a dyadic conversation: ``what you are replying to", ``who you are" and ``who you are talking to". ``What you are replying to" is the previous message of the replying sentence, indicating the cause of the corresponding response. ``Who you are" and ``who you are talking to" are personal information of both sides of the conversation, indicating how an individual tends to talk and be replied to. Thus, we propose an augmented GRU model via sequence to sequence learning and adjusting model formulation with personal information. With the learned PCE of each individual, we can make personality inferences based on two-layer fully-connected neural networks.

To the best of our knowledge, we are the first to predict user personality in a conversational scenario. The main contributions of this work are listed as follows:
\begin{itemize}
\item
We address the challenge of personality inference with dyadic conversations. This can utilize various contextual information in conversation records for better inferring personality of individuals.

\item
We propose an augmented GRU model for learning unsupervised personal conversational embeddings based on dyadic conversation. Besides modeling sentences with GRU, our method can model several important contextual factors in dyadic conversation.

\item
We collect and conduct experiments on the Movie Script dataset. We find modeling dyadic conversations can significantly improve the accuracy of personality inference, and experimental results show the strong performance of our method.

\end{itemize}

The rest of the paper is organized as follows. In section 2, we review some related works on personality inference and recurrent neural networks. Section 3 details our method for learning personal conversational embeddings based on dyadic conversation. In section 4, we report and analyze experimental results. Section 5 concludes our work and discusses future research.

\begin{figure}[tb]
\centering
\includegraphics[width=0.5\textwidth]{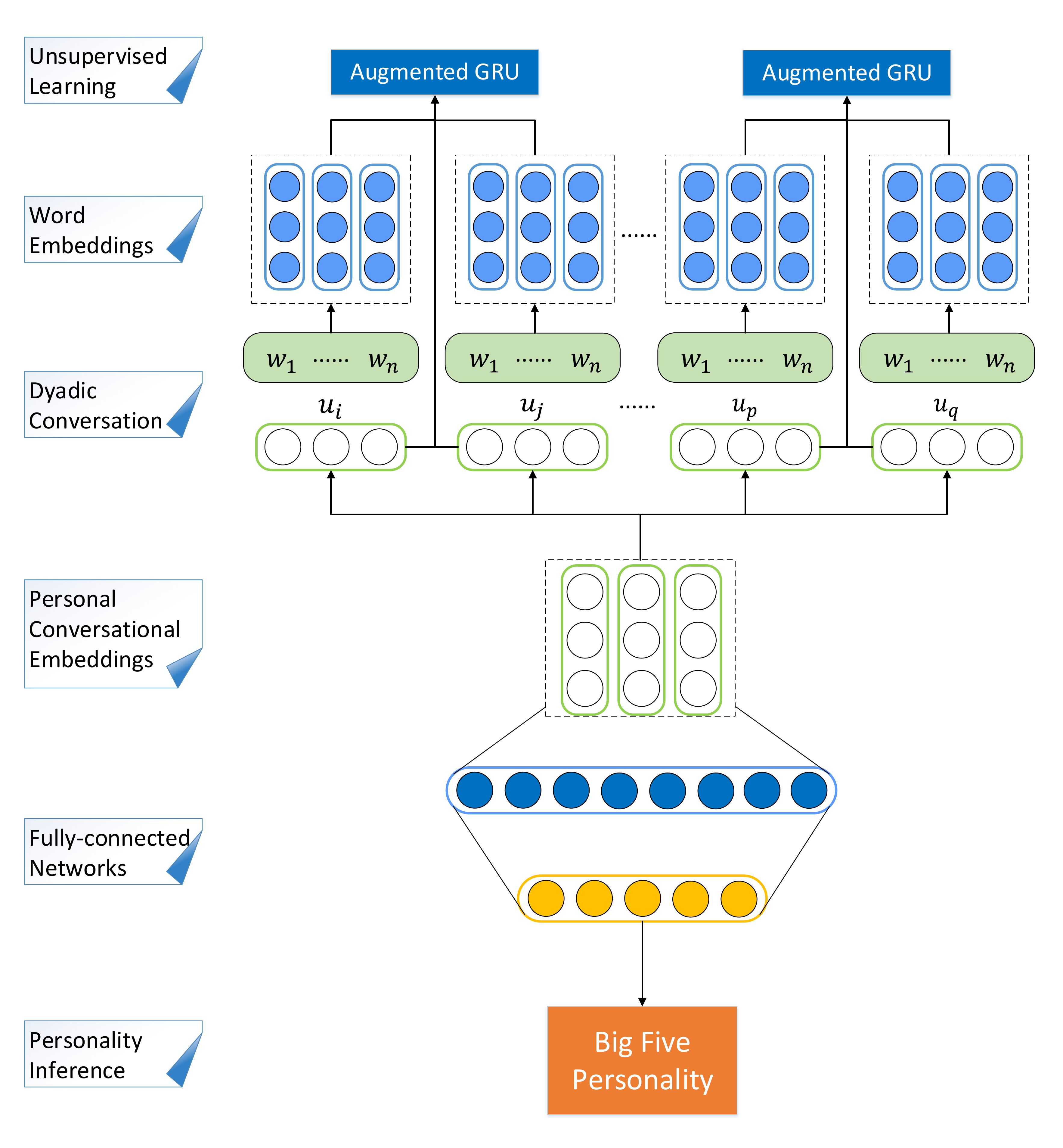}
\caption{Framework of learning Personal Conversational Embeddings (PCE) and inferring the big five personality based on dyadic conversations between individuals.}
\label{fig:framework}
\end{figure}

\section{Related Work}

In this section, we briefly review some related work on personality inference and recurrent neural networks.

\subsection{Personality Inference}

Nowadays, social media provides a great amount of personal information on users, e.g., social relationships, tweets, blogs and uploaded photos. The personality of an individual can be extracted from these user generated contents \cite{golbeck2011predicting}, and extensive research has been done on automatic personality inference \cite{farnadi2016computational}. Facebook\footnote{http://www.facebook.com} has collected data on users and started a project called myPersonality\footnote{http://mypersonality.org}, which has attracted significant attention from researchers \cite{bachrach2012personality,farnadi2013recognising,hagger2011social,kosinski2013private,schwartz2013personality,youyou2015computer}. Different types of features have been investigated on Facebook data, such as user profiles \cite{hagger2011social}, using patterns \cite{bachrach2012personality}, likes \cite{kosinski2013private,youyou2015computer} and textual contents \cite{farnadi2013recognising,schwartz2013personality}. Bachrach et al. \cite{bachrach2012personality} analyze the correlation between Facebook using patterns and users' personality. Wu et al. \cite{youyou2015computer} apply matrix factorization on user-item matrix for personality inference, and conclude that computer-based automatic personality inference is more accurate than those made by humans. Schwartz et al. \cite{schwartz2013personality} infer users' personality, gender, and age by applying topic model on user posted contents. Data on Twitter has also been utilized for analyzing personality \cite{golbeck2011twitter,hughes2012tale,liu2016analyzing,quercia2011our}. Hughes et al. \cite{hughes2012tale} investigate the difference between Facebook usage and Twitter usage for analyzing personality. Quercia et al. \cite{quercia2011our} predict personality based user profiles, such as followings, followers, and listed counts. Liu et al. \cite{liu2016analyzing} predict personality via extracting features from users' avatars and uploaded photos.
Meanwhile, similar technologies have also bee used for verifying misinformation on social media \cite{wu2016information,ma2016detecting,yu2017convolutional,liu2018mining,yu2019attention}.

\subsection{Recurrent Neural Networks}

Recently, RNN has been widely applied for various sentence modeling tasks. Back-Propagation Through Time (BPTT) \cite{rumelhart1988learning} is usually used for optimization of RNN models. However, in practice, basic RNN structure will face the vanishing or exploding gradients problem when learning long-term temporal dependency \cite{bengio1994learning}. To overcome this drawback, researchers extend RNN with memory units, and propose some advanced structures, such as LSTM \cite{hochreiter1997long} and GRU \cite{chung2014empirical,chung2015gated}. According to existing research \cite{chung2014empirical}, LSTM and GRU share similar performances on various tasks, but GRU has much lower computational cost.

Recently, RNN, including LSTM and GRU, has been widely applied for generating conversation and response \cite{li2016persona,serban2016building,shang2015neural,sordoni2015neural,vinyals2015neural}. Basically, this is a sequence-to-sequence problem. Vinyals et al. \cite{vinyals2015neural} build the first neural conversational model based on RNN. Shang et al. \cite{shang2015neural} incorporate an encoder-decoder framework in GRU for generating responses. Serban et al. \cite{serban2016building} propose a hierarchical model for conversation machines. Sordoni et al. \cite{sordoni2015neural} utilize previous paragraphs as contexts, and propose a context-sensitive response generating model. Li et al. \cite{li2016persona} adjust LSTM formulation with personal information, such as address and birthday, to better generate personalized conversational response.
These methods have also been applied in various sequential prediction \cite{yu2016dream,wang2019towards,hidasi2016session,zhang2014sequential}.

Variety of contextual information has also been incorporated in RNN models for different specific tasks. Ghosh et al. \cite{ghosh2016contextual} apply topic model and treat surrounding texts of a specific sentence as contexts. Then each layer of LSTM is adjusted with contexts, and the Contextual LSTM (CLSTM) model is proposed. CLSTM achieves state-of-the-art performance in next sentence selection and sentence topic prediction. Visual features have also been incorporated in LSTM for multimodal applications, such as image caption \cite{mao2014deep} and visual question answering \cite{ren2015exploring}. Moreover, RNN has been jointly modeled with behavioral contexts for user modeling, and achieves state-of-the-art performances in recommender systems \cite{wu2016sape,liu2016context,liu2016strnn,wu2017context,liu2017multi,cui2018mv}.

\section{Modeling Dyadic Conversation}

In this section, we present our method for learning unsupervised personal conversational embeddings based on dyadic conversation. First, we formulate the notations of this paper. Then, we introduce conventional RNN and GRU structures. Finally, we discuss how to involve several contextual factors, i.e., ``who you are", ``what you are replying to" and ``who you are talking to", in dyadic conversation into an augmented GRU model.

\subsection{Notations}

In this work, we have a set of individuals denoted as $U$$=$$\{$$u_1,$$u_2,$$...$$\}$, and conversation records between them. Each pair of dyadic conversation between two individuals consists of a message (the sentence before the corresponding response) and a response (the replying sentence) denoted as $M$$=$$\{$$BOS,$$w_1,$$w_2,$$...,$$EOS$$\}$ and $R$$=$$\{$$BOS,$$w_1,$$w_2,$$...,$$EOS$$\}$ respectively. $BOS$ and $EOS$ are begin and end of the word sequence respectively, and $w_t$ means one word in a sentence (message or response). Sentences in the conversation are associated with personal information of both sides of the conversation. Given conversation records of individuals, we plan to infer each individual's big five personality: extraversion, agreeableness, conscientiousness, neuroticism, and openness.

\subsection{Recurrent Neural Networks}
The architecture of RNN is a recurrent structure with multiple hidden layers. At each time step, we can predict the output unit given the hidden layer, and then feed the new output back into the next hidden status. It has been successfully applied in a variety of applications \cite{ghosh2016contextual,liu2016context,mao2014deep,vinyals2015neural}. The formulation of each hidden layer in RNN is:
\begin{equation}  \label{RNN}
{{\mathbf{h}}_t} = tanh({{\mathbf{w}}_t}{\mathbf{M}} + {{\mathbf{h}}_{t - 1}}{\mathbf{N}})~,
\end{equation}
where $\mathbf{w}_t \in {\mathbb{R}^{d}}$ is the word embedding, $\mathbf{h}_t \in {\mathbb{R}^{d}}$ is the hidden state of RNN, and ${\mathbf{M}} \in {\mathbb{R}^{d \times d}}$ and ${\mathbf{N}} \in {\mathbb{R}^{d \times d}}$ are transition matrices in RNN.

\subsection{Gated Recurrent Units}

Due to the vanishing or exploding gradients problem \cite{bengio1994learning}, it is hard for conventional RNN to learn long-term dependency in sequences. Accordingly, memory units are proposed and applied, e.g., LSTM \cite{hochreiter1997long} and GRU \cite{chung2014empirical,chung2015gated}. LSTM and GRU share similar performances in various tasks, but GRU is much faster \cite{chung2014empirical}. So, GRU is becoming a better method for learning embeddings in sentences. The formulation of each layer of GRU is:
\begin{equation}\label{GRU}
\begin{array}{l}
{{\mathbf{z}}_t} = \sigma \left( {{{\mathbf{w}}_t}{{\mathbf{M}}_z} + {{\mathbf{h}}_{t - 1}}{{\mathbf{N}}_z}} \right)\\
{{\mathbf{r}}_t} = \sigma \left( {{{\mathbf{w}}_t}{{\mathbf{M}}_r} + {{\mathbf{h}}_{t - 1}}{{\mathbf{N}}_r}} \right)\\
{{{\mathbf{\tilde h}}}_t} = tanh\left( {{{\mathbf{w}}_t}{{\mathbf{M}}_h} + \left( {{{\mathbf{h}}_{t - 1}} \cdot {{\mathbf{r}}_t}} \right){{\mathbf{N}}_h}} \right)\\
{{\mathbf{h}}_t} = \left( {1 - {{\mathbf{z}}_t}} \right) \cdot {{\mathbf{h}}_{t - 1}} + {{\mathbf{z}}_t} \cdot {{{\mathbf{\tilde h}}}_t}
\end{array}~,
\end{equation}
where $\mathbf{w}_t \in {\mathbb{R}^{d}}$ is the embedding of a word in a sentence, and $\mathbf{h}_t \in {\mathbb{R}^{d}}$ is the hidden state of GRU. ${{\mathbf{z}}_t}$ is a reset gate, determining how to combine the new input with the previous memory. ${{\mathbf{r}}_t}$ is an update gate, defining how much of the previous memory is cascaded into the current state. ${{{\mathbf{\tilde h}}}_t}$ denotes the candidate activation of the hidden state ${{\mathbf{h}}_t}$.

\begin{figure*}[tb]
\centering
\includegraphics[width=1\textwidth]{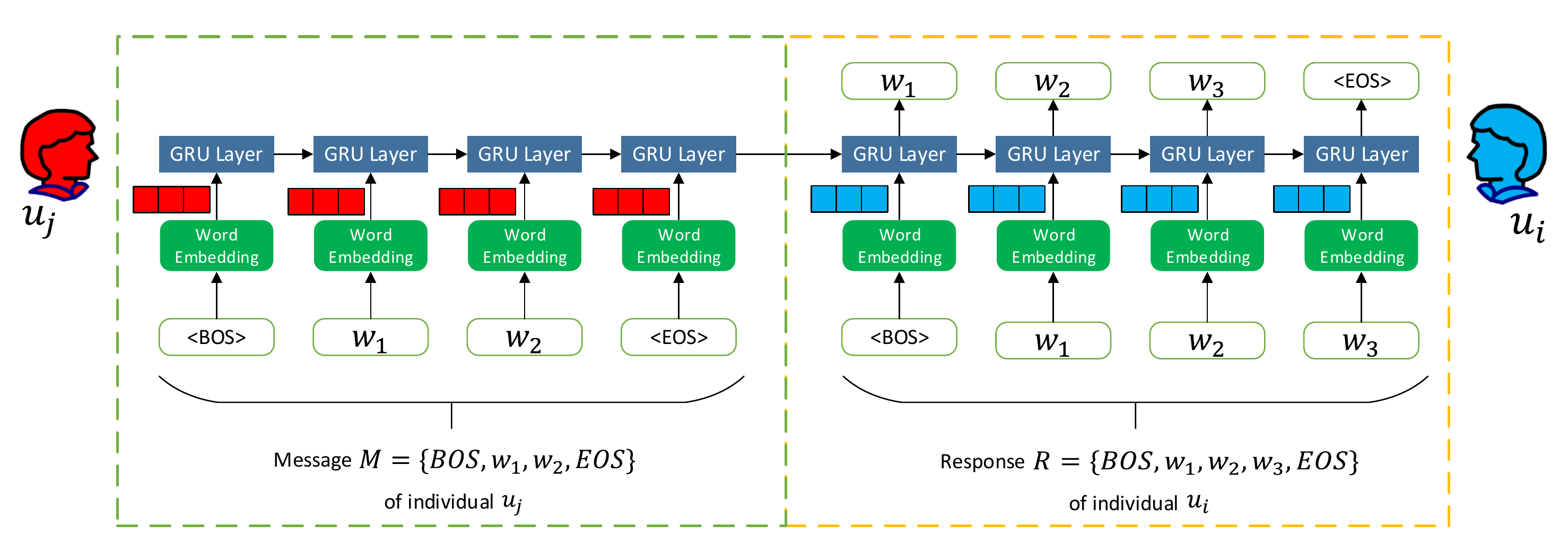}
\caption{The sequence to sequence learning process of dyadic conversation between $u_i$ and $u_j$.}
\label{fig:s2s}
\end{figure*}

\subsection{Personal Speaking Embeddings}

Though GRU has succeeded in modeling sentences, it can not be directly applied for personality analysis. GRU is not able to involve the personal information of the current speaker into consideration. However, personal information, including word preferences and language usage patterns, is vital for personality inference. Research in conversation machine has done some investigation on modeling personal information \cite{li2016persona}. Thus, we incorporate personal information with the conventional GRU model, and learn \textbf{Personal Speaking Embeddings (PSE)} based on sentences generated by individuals. The learning procedure is unsupervised along with the structure of GRU via predicting the next word, which makes our method flexible and does not require much annotated data. The PSE models ``who you are", capturing the speaker's language usage preferences reflected in sentences in the conversation. Mathematically, incorporating the personal information of the current speaking individual $u_i$, the formulation of each layer of GRU becomes:
\begin{equation}\label{PSE}
\begin{array}{l}
{{\mathbf{z}}_t} = \sigma \left( {{{\mathbf{w}}_t}{{\mathbf{M}}_z} + {{\mathbf{h}}_{t - 1}}{{\mathbf{N}}_z}{\rm{ + }}{{\mathbf{u}}_i}{{\mathbf{P}}_z}} \right)\\
{{\mathbf{r}}_t} = \sigma \left( {{{\mathbf{w}}_t}{{\mathbf{M}}_r} + {{\mathbf{h}}_{t - 1}}{{\mathbf{N}}_r}{\rm{ + }}{{\mathbf{u}}_i}{{\mathbf{P}}_r}} \right)\\
{{{\mathbf{\tilde h}}}_t} = tanh\left( {{{\mathbf{w}}_t}{{\mathbf{M}}_h} + \left( {{{\mathbf{h}}_{t - 1}} \cdot {{\mathbf{r}}_t}} \right){{\mathbf{N}}_h}{\rm{ + }}{{\mathbf{u}}_i}{{\mathbf{P}}_h}} \right)\\
{{\mathbf{h}}_t} = \left( {1 - {{\mathbf{z}}_t}} \right) \cdot {{\mathbf{h}}_{t - 1}} + {{\mathbf{z}}_t} \cdot {{{\mathbf{\tilde h}}}_t}
\end{array}~,
\end{equation}
where ${\mathbf{u}}_i \in {\mathbb{R}^{d}}$ is the embedding of current speaking individual, which can be learned unsupervisedly in the GRU structure. Formulation of the reset gate, the update gate, and the hidden state candidate activation are adjusted with personal information via the calculation ${{\mathbf{u}}_i}{{\mathbf{P}}_z}$, ${{\mathbf{u}}_i}{{\mathbf{P}}_r}$ and ${{\mathbf{u}}_i}{{\mathbf{P}}_h}$ respectively. The learned embedding ${\mathbf{u}}_i$ captures how the individual prefers to speak and express himself or herself. This process can compress all the sentences in the conversation generated by an individual into a fixed dimensional latent representation. With the compressed representation of each individual, personality analysis and inference can be performed.

\subsection{Personal Replying Embeddings}

There is another important contextual factor in dyadic conversation: the message before the response, i.e., the sentence before the corresponding replying sentence. This refers to ``what you are replying to", capturing the cause and reason of the replying sentence. It is vital for analyzing personality. For example, if someone says something impolite to you, you may also say some rude words. But this does not mean you are a rude person. Thus, we incorporate the message before the response in conversation into our model via sequence to sequence learning, and learn \textbf{Personal Replying Embeddings (PRE)}. The sequence to sequence learning is a widely-used structure for modeling message and response in question answering and conversation machine \cite{li2016persona,serban2016building,shang2015neural,sordoni2015neural,vinyals2015neural}.

Figure \ref{fig:s2s} illustrates the sequence to sequence learning process of dyadic conversation between $u_i$ and $u_j$. The model is usually learned with an encoder-decoder method. For message $M$$=$$\{$$BOS,$$w_1,$$w_2,$$...,$$EOS$$\}$ generated by individual $u_j$, the sentence is encoded into a sentence embedding with the GRU model and fed to the following decoder as input. For response $R$$=$$\{$$BOS,$$w_1,$$w_2,$$...,$$EOS$$\}$ generated by individual $u_i$, the sentence is modeled and decoded according to the formulation in Equation \ref{PSE}. The learned PRE takes causes of replying sentences into consideration, reducing noise in data and being more flexible. To the best of our knowledge, no existing personality inference methods can model the message before the replying sentence in conversational data.

\begin{figure}[tb]
\centering
\includegraphics[width=0.5\textwidth]{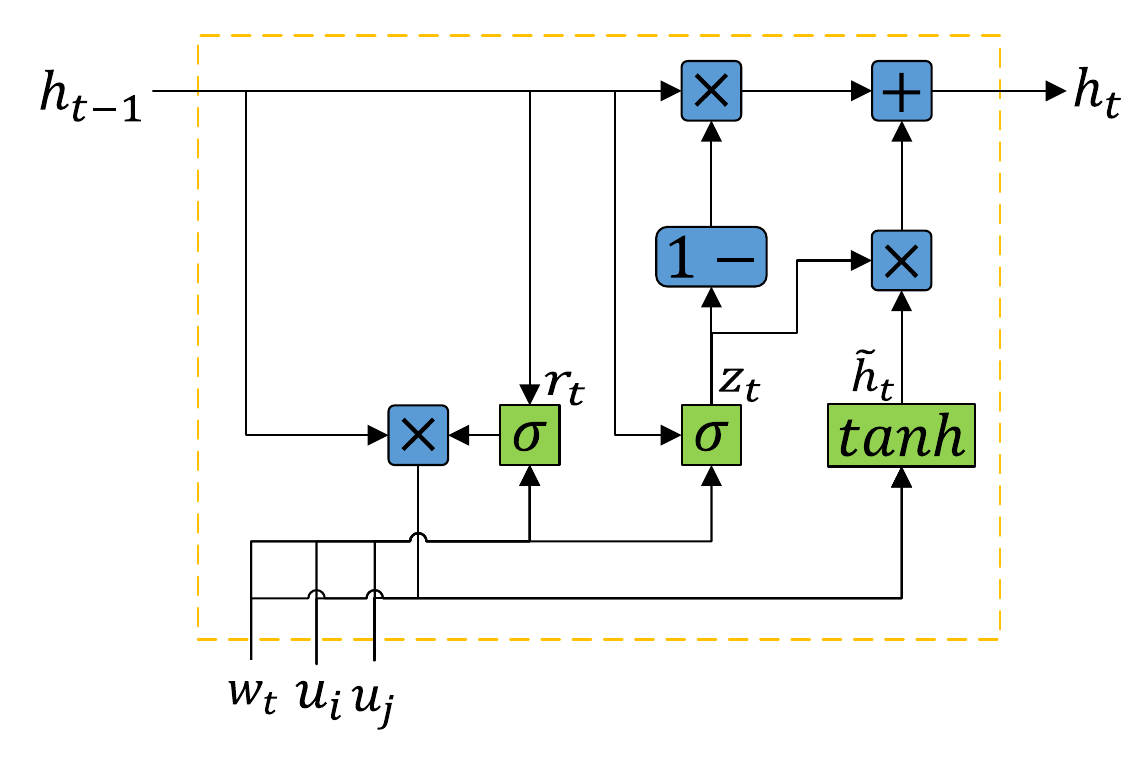}
\caption{Structure of the augmented GRU model for learning Personal Conversational Embeddings (PCE) incorporating the personal information of both sides of the dyadic conversation.}
\label{fig:geu}
\end{figure}

\subsection{Personal Conversational Embeddings}

\begin{table*}[tb]
  \centering
  \caption{Details of the Movie Script dataset.}
    \begin{tabular}{ccccc}
    \hline
    dataset & \#individuals & \#sentences & language & description \\
    \hline
    Movie Script & 880   & 180k  & English & conversation between characters in movie scripts \\
    \hline
    \end{tabular}%
  \label{tab:data}%
\end{table*}%

Furthermore, the personal information of the individual being replied to is also useful for analyzing personality. It tells us ``who you are talking to", and captures how an individual tends to be replied to. For example, a friendly person usually receives polite treatment, while a rude person usually receives unfriendly words. Accordingly, we incorporate the personal information of the individual being replied to in our model, and learn \textbf{Personal Conversational Embeddings (PCE)} based on all the contextual information in dyadic conversation. For the conversation between $u_i$ and $u_j$, the formulation in Equation \ref{PSE} can be adjusted as:
\begin{equation}\label{PCE}
\begin{array}{l}
{{\mathbf{z}}_t} = \sigma \left( {{{\mathbf{w}}_t}{{\mathbf{M}}_z} + {{\mathbf{h}}_{t - 1}}{{\mathbf{N}}_z}{\rm{ + }}{{\mathbf{u}}_i}{{\mathbf{P}}_z} + {{\mathbf{u}}_j}{{\mathbf{Q}}_z}} \right)\\
{{\mathbf{r}}_t} = \sigma \left( {{{\mathbf{w}}_t}{{\mathbf{M}}_r} + {{\mathbf{h}}_{t - 1}}{{\mathbf{N}}_r}{\rm{ + }}{{\mathbf{u}}_i}{{\mathbf{P}}_r} + {{\mathbf{u}}_j}{{\mathbf{Q}}_r}} \right)\\
{{{\mathbf{\tilde h}}}_t} = tanh\left( {{{\mathbf{w}}_t}{{\mathbf{M}}_h} + \left( {{{\mathbf{h}}_{t - 1}} \cdot {{\mathbf{r}}_t}} \right){{\mathbf{N}}_h}{\rm{ + }}{{\mathbf{u}}_i}{{\mathbf{P}}_h} + {{\mathbf{u}}_j}{{\mathbf{Q}}_h}} \right)\\
{{\mathbf{h}}_t} = \left( {1 - {{\mathbf{z}}_t}} \right) \cdot {{\mathbf{h}}_{t - 1}} + {{\mathbf{z}}_t} \cdot {{{\mathbf{\tilde h}}}_t}
\end{array}~,
\end{equation}
where ${\mathbf{u}}_j \in {\mathbb{R}^{d}}$ is the embedding of the individual being replied to. To be noted that, embeddings of current speaking individuals and embeddings of individuals being replied to share same parameters. Formulation of the reset gate, the update gate and the hidden state candidate activation can be adjusted via the calculation ${{\mathbf{u}}_j}{{\mathbf{Q}}_z}$, ${{\mathbf{u}}_j}{{\mathbf{Q}}_r}$ and ${{\mathbf{u}}_j}{{\mathbf{Q}}_h}$ respectively. The learned PCE captures how an individual tends to speak and be replied to, as well as reasons and causes of responses in conversation. It can compress language usage and all the contextual information in the dyadic conversation into a latent personal embedding, which presents perspectives for user modeling and personality analysis.

\subsection{Learning and Prediction}

Our GRU models, including word embeddings and personal embeddings, can be learned via predicting the next word in the sentence with a cross-entropy loss. We train all models by employing the derivative of the loss with respect to all parameters through the BPTT \cite{rumelhart1988learning} algorithm. We iterate over all the samples in the conversational data in each epoch until the convergence is achieved or a maximum epoch number is met. We empirically set the vocabulary size as $5000$, the learning rate as $0.01$, the maximum epoch number as $100$, and select the dimensionality of hidden embeddings in $[25,50,75,100]$.

After personal embeddings, including PSE, PRE and PCE, are well learned with the GRU model, prediction of personality inference can be made. Here, we apply two-layer fully-connected neural networks. In the first layer, we map embeddings from the conversational space to the personality space. For embedding $\mathbf{u}_i$, the formulation can be:
\begin{equation}\label{p1}
{{\mathbf{s}}_i} = \sigma \left( {{{\mathbf{u}}_i}{{\mathbf{W}}_1} + {{\mathbf{c}}_1}} \right)~,
\end{equation}
where ${\mathbf{s}}_i \in {\mathbb{R}^{d'}}$, ${\mathbf{W}}_1 \in {\mathbb{R}^{d \times d'}}$ and ${\mathbf{c}}_1 \in {\mathbb{R}^{d'}}$. ${\mathbf{s}}_i$ is the joint representation of five traits of the big five personality in a common space. For simplification, we select dimensionality as $d'=d$ in this paper. Then, in the second layer, we can make predictions on the big five personality traits:
\begin{equation}\label{p2}
{{\mathbf{y}}_i} = f\left( {{{\mathbf{s}}_i}{{\mathbf{W}}_2} + {{\mathbf{c}}_2}} \right)~,
\end{equation}
where ${\mathbf{y}}_i \in {\mathbb{R}^{5}}$, ${\mathbf{W}}_2 \in {\mathbb{R}^{d' \times 5}}$, ${\mathbf{c}}_2 \in {\mathbb{R}^{5}}$, and $f(x)$ is chosen as a $sigmoid$ function $f(x) = \exp \left( {1 \mathord{\left/ \right. \kern-\nulldelimiterspace} 1 + e^{ - x} } \right)$. ${\mathbf{y}}_i$ is the predicted values of the five personality traits: extraversion, agreeableness, conscientiousness, neuroticism, and openness. The prediction model can be trained with the widely-used back-propagation algorithm.

\section{Experiments}

In this section, we introduce our experiments. First, we introduce our experimental dataset, i.e., the Movie Script dataset, and several compared methods. Then, we give comparison among different methods. We also investigate the dimensionality impact and the consistency of personal embeddings. Finally, we apply the learned personal conversational embeddings in individual retrieval and illustrate some results on the Movie Script dataset.

\subsection{Data}

To investigate the effectiveness of personal conversational embeddings, we collect the Movie Script dataset.

\textbf{The Movie Script dataset} is collected via crawling conversations between characters in movie scripts from Internet Movie Script Database (IMSDB)\footnote{http://www.imsdb.com/}. IMSDB is a website containing thousands of English movie scripts all over the world. Among them, we select 200 movies according to their popularity. The popularity is determined based on the number of ratings on IMDB\footnote{http://www.imdb.com/} of each movie, where the threshold is set as 100k. Data from movie scripts has been widely used for constructing datasets for question answering tasks \cite{li2016persona}. Moreover, as we know, the personality of characters is important in scripts and novels. In a good script or novel, the personality of a character can cause his or her actions and promote the development of the plot. Thus, it is reasonable to utilize movie scripts for our experiments. The dataset contains a total of about 180k sentences. To avoid data sparsity, we select about 880 characters with at least 50 sentences in scripts. We manually annotate these characters with the big five personality traits for training and evaluation. We label each personality trait in a binary value.

\begin{figure*}[tb]
\centering
\includegraphics[width=0.98\textwidth]{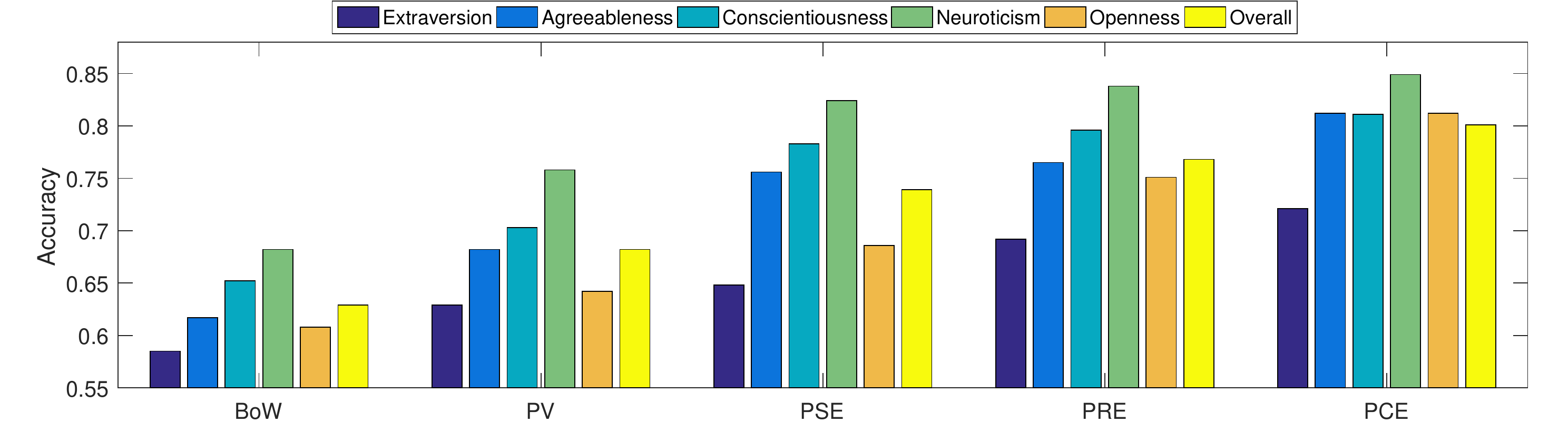}
\caption{Performance comparison on inferring the big five personality traits evaluated by accuracy. The dimensionality of embeddings is set to be $d=50$. The larger the value, the better the performance.}
\label{performance comparison}
\end{figure*}

Details of the Movie Script dataset are illustrated in Table \ref{tab:data}. We randomly select $80\%$ of individuals for training, and the remaining $20\%$ for testing.

\subsection{Compared Methods}

In our experiments, the following methods are implemented and compared on above two datasets:

\begin{itemize}

\item
\textbf{Bag of Words (BoW)}. BoW is a simple baseline method in our experiments. BoW features are extracted on all the sentences an individual has said in the dataset. This method does not utilize the contextual information in conversation. Due to the limitation of annotated data, we select the $300$ most significant words according to Pearson correlation coefficient\footnote{https://en.wikipedia.org/wiki/Pearson\_product-moment\_correlation\_coefficient}. Based on the $300$-dimensional feature, softmax is performed for personality inference.

\item
\textbf{Paragraph Vector (PV)} \cite{le2014distributed}. PV is a state-of-the-art method to learn unsupervised embeddings for paragraphs based on word co-occurrence. We treat all contents generated by an individual as a paragraph, on which PV features are extracted. This method can compress all the contents of an individual into a fixed-dimensional feature. Then, softmax is performed for personality inference. Obviously, this is a method that lacks modeling of the conversational information.

\item
\textbf{Personal Speaking Embeddings (PSE)}. Based on an augmented GRU model, PSE learns personal embeddings based on conversational data. This is a baseline among our three proposed methods.

\item
\textbf{Personal Replying Embeddings (PRE)}. Extended from PSE, PRE incorporates an important contextual factor in dyadic conversation: the message before the corresponding replying sentence. This can explain the cause of the replying sentence and reduce noise in the data.

\item
\textbf{Personal Conversational Embeddings (PCE)}. As the most advanced one among our methods, PCE models all contextual information in dyadic conversation: the message before the response, as well as the personal information of both sides of the conversation. It captures how an individual tend to speak and be replied to, as well as reasons and causes of responses in conversation. It can compress language usage and all contextual information in the dyadic conversation into a latent personal embedding.

\end{itemize}

\begin{table*}[tb]
  \centering
  \caption{Some results of individual retrieval with personal conversational embeddings on the Movie Script dataset. We illustrate three most similar individuals of each query in the table.}
    \begin{tabular}{cccc}
    \hline
    query individual & most similar individuals & query individual & most similar individuals \\
    \hline
    \emph{Marlin} & \emph{Carl} in ``Up" & \emph{Red}   & \emph{Master Oogway} in ``Kung Fu Panda" \\
    in    & \emph{Gru} in ``Despicable Me" & in    & \emph{Lincoln} in ``Lincoln" \\
    ``Finding Nemo" & \emph{Po's dad} in ``Kung Fu Panda" & ``Shawshank Redemption" & \emph{Chuck} in ``Cast Away" \\
    \hline
    \emph{Dory}  & \emph{Po} in ``Kung Fu Panda" & \emph{Tyler} & \emph{Jordan} in ``The Wolf of Wall Street" \\
    in    & \emph{Olif} in ``Frozen" & in    & \emph{Holmes} in ``Sherlock Holmes" \\
    ``Finding Nemo" & \emph{Agnes} in ``Despicable Me" & ``Fight Club" & \emph{Calvin} in ``Django Unchained" \\
    \hline
    \emph{Michael} & \emph{Michael} in ``The Godfather 2" & \emph{Fletcher} & \emph{Bruce} in ``Bruce Almighty" \\
    in    & \emph{Vito} in ``The Godfather 2" & in    & \emph{Ace} in ``Ace Ventura: Pet Detective" \\
    ``The Godfather" & \emph{Michael} in ``The Godfather 3" & ``Liar Liar" & \emph{Stanley} in ``The Mask" \\
    \hline
    \end{tabular}%
  \label{tab:retrieval}%
\end{table*}%

\subsection{Performance Comparison}

To investigate the effectiveness of our proposed methods incorporating contextual information in dyadic conversation, we illustrate performance comparison with personality $d=50$ on the Movie Script dataset in Figure \ref{performance comparison}. Results are evaluated by accuracies of the big five personality traits, i.e., extraversion, agreeableness, conscientiousness, neuroticism, and openness, as well as the overall accuracy of the big five personality.

According to results on the Movie Script dataset in Figure \ref{performance comparison}, PV outperforms BoW, indicating that a compressed embedding is better than directly analyzing bag of words for inferring personality. Constructed based on the GRU model, PSE can further improve the performance of PV, showing the advantage of RNN models for learning personal embeddings. Furthermore, modeling the message before the corresponding replying sentence, PRE outperforms PSE with a relatively large advantage. PCE, which can incorporate the personal information of both side of the dyadic conversation, has better performance comparing with PRE, and achieves best accuracies on all five personality traits. PCE relatively improves the overall accuracy by $27.3\%$, $17.4\%$, $8.4\%$ and $4.3\%$ comparing with BoW, PV, PSE and PRE respectively. Meanwhile, we observe that, neuroticism has the highest inference accuracy among five personality traits. This may indicate that, whether an individual is sensitive, angry and depressed is distinctive in dramatic plots.
These significant improvements shown in figures indicate the superiority of our method brought by modeling the rich contextual information in the dyadic conversation with an augmented GRU model.

\subsection{Impact of Dimensionality}

\begin{figure}[tb]
\centering
\includegraphics[width=0.48\textwidth]{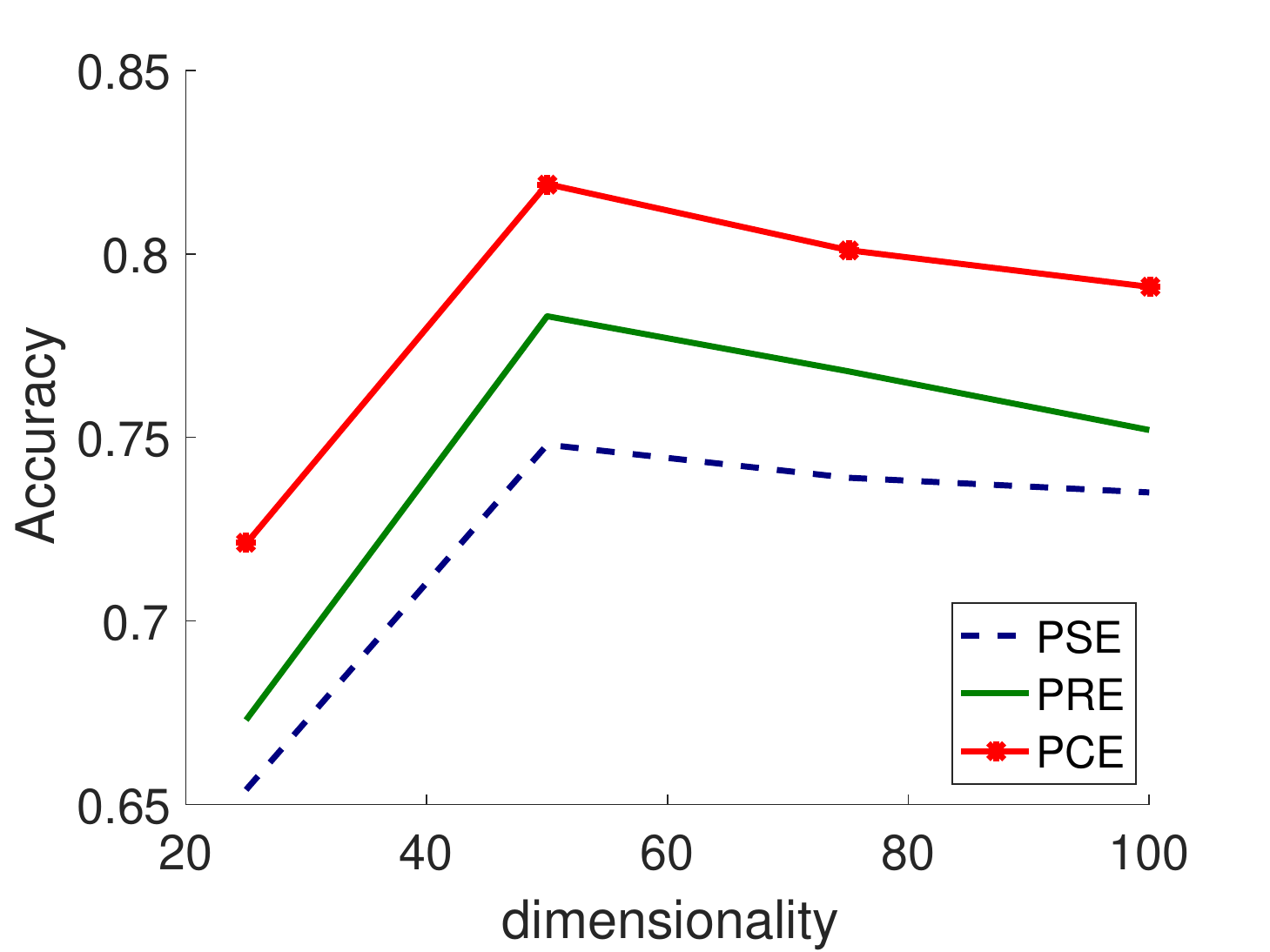}
\caption{Performance of personal embeddings with varying dimensionality $d=[25,50,75,100]$. Performances are evaluated by the overall accuracy of inferring the big five personality traits.}
\label{dimen}
\end{figure}

To investigate the impact of dimensionality on learning personal embeddings based on dyadic conversation, and select the best parameter for personality inference, we illustrate the performance of PSE, PRE and PCE with varying dimensionality $d=[25,50,75,100]$ in Figure \ref{dimen}. Results are evaluated by the overall accuracy of inferring five traits in the big five personality.

From results in Figure \ref{dimen}, we can obtain some observations. First, we can observe similar performance comparison among PSE, PRE and PCE as in the previous subsection. Second, the best dimensionality of learning personal embeddings is clearly $d=50$, which is used in the rest of our experiments. Third, performances are stable with a large range of dimensionality in $d=[50,75,100]$, indicating the flexibility of our methods. Forth, PSE, PRE and PCE share very similar tendency along with dimensionality. But methods incorporating more contextual information tend to be more easy for overfitting the data when dimensionality is high, where the performance of PCE is even worse than that of PRE when $d>50$.

\subsection{Consistency of Embeddings}

Despite having good performance in personality inference, the embedding of an individual should consistently represent the individual, and should not change very much in different time periods and situations. Accordingly, we should conduct experiments to evaluate the consistency of personal embeddings. Thus, we divide data from an individual in the dataset into two equal parts in chronological order. For two part of an individual's conversational data, embeddings are learned and features are extracted separately. To avoid data sparsity, we only select individuals with at least $100$ sentences. Experimental results on the Movie Script dataset are shown in Table \ref{tab:consistency}.

First, we use the first part of an individual to retrieve for the second part of the same individual based on Euclidean distance. With better embeddings, more correct pairs should be matched. The evaluation metrics used here is recall. Recall is calculated as the ratio of the number of correctly found pairs and the total number of positive samples. The larger the recall value, the better the consistency. From results in Table \ref{tab:consistency}, we can observe that PRE and PCE can significantly outperform other methods on both datasets. On the Movie Script dataset, the recall value of PCE is slightly better than that of PRE.

Second, we evaluate the difference between embeddings of two parts of the same individual. We use Root Mean Square Error (RMSE)\footnote{https://www.kaggle.com/wiki/RootMeanSquaredError} to evaluate embedding differences. The smaller the RMSE value, the better the consistency. Because BoW features are not real value embeddings, and PV is not trained with GRU, they have different ranges of values comparing with PSE, PRE and PCE. It is not meaningful to compute their embedding differences to compare with our proposed personal embeddings. So, we ignore BoW and PV here. Results in Table \ref{tab:consistency} show the lowest RMSE values of PCE with relatively large advantages. These results indicate the good consistency of personal conversational embeddings for representing individuals.

\begin{table}[tb]
  \centering
  \caption{Evaluation of the consistency of personal embeddings via comparing two parts of the conversation of the same individual. Results are evaluated by recall and RMSE.}
    \begin{tabular}{ccc}
    \hline
    \multirow{2}[0]{*}{} & \multicolumn{2}{c}{Movie Script} \\
          & recall & RMSE \\
    \hline
    BoW   & 0.198 & ------ \\
    PV    & 0.256 & ------ \\
    PSE   & 0.307 & 0.067 \\
    PRE   & 0.419 & 0.043 \\
    PCE   & \textbf{0.440} & \textbf{0.035} \\
    \hline
    \end{tabular}%
  \label{tab:consistency}%
\end{table}%

\subsection{Individual Retrieval}

In this subsection, we investigate a case study on individual retrieval on the Movie Script dataset to demonstrate advantages of personal conversational embeddings. We use some famous characters in movies as queries, and retrieve for other similar characters in the dataset based on Euclidean distance with personal conversational embeddings learned from their conversations. We illustrate some representative results in Table \ref{tab:retrieval}. For each query, three most similar characters are shown.

From the illustration, we can observe some interesting phenomena and suitable matchings. In the animated film ``Finding Nemo", we have two characters, \emph{Marlin} and \emph{Dory}. \emph{Marlin} is matched with \emph{Carl}, \emph{Gru}, and {Po's dad}, while \emph{Dory} is matched with \emph{Po}, \emph{Olif}, and \emph{Agnes}. Former ones are conservative fathers, and latter ones are childish characters. \emph{Red} in ``Shawshank Redemption" is matched with \emph{Master Oogway}, \emph{Lincoln}, and \emph{Chuck}, who are all wise elders. \emph{Tyler} in ``Fight Club" is matched with \emph{Jordan}, \emph{Holmes}, and \emph{Calvin}, who are all insane, insolent, and arrogant individuals. \emph{Michael} in ``The Godfather" is matched with himself in other chapters of the movie series, as well as his father at a young age. Due to the consistent nonsensical style of Jim Carrey's comedies, several roles he played, i.e., \emph{Fletcher}, \emph{Bruce}, \emph{Ace} and \emph{Stanley}, are matched. Obviously, matched characters share very similar personalities. This indicates that, personal conversational embeddings are able to well capture the characteristics of an individual.

\section{Conclusions and Future Work}

In this paper, we propose a novel method for personality inference. With an augmented GRU model, we learn unsupervised personal conversational embeddings based on dyadic conversation between individuals. We incorporate previous sentences of replying sentences via sequence to sequence learning. The formulation of each layer of networks is adjusted with personal information of both sides of the dyadic conversation. With the learned personal conversational embeddings, personality inference can be performed. According to experiments conducted on the Movie Script dataset, accuracy of personality inference can be significantly improved by modeling the rich contextual information in conversation records. Experimental results show the successful performance of our proposed method.

In the future, we will further investigate the following directions. First, social relationship is an important factor for personality analysis. So, it is reasonable to investigating the relationship between individuals in the conversation. Second, we can incorporate more contextual information, such as location, time, recent topics or even social environment. The main challenge for doing this is finding available data.


%

%
%

\ifCLASSOPTIONcaptionsoff
  \newpage
\fi



%
\balance
\bibliographystyle{IEEEtran}
\bibliography{PCE}

%

%
%
%




\end{document}